\documentclass{article}
\usepackage{arxiv}

\pdfoutput=1

\usepackage{amsmath}
\usepackage{graphicx,changepage}

\usepackage{framed}
\usepackage[hidelinks]{hyperref}
\usepackage{natbib}     
\usepackage{graphics}
\usepackage{listings}	
\usepackage{multicol}

\usepackage{xcolor}
\lstdefinelanguage{PDDL}
{
    showlines=true,
	basicstyle=\small\ttfamily,
	sensitive=false,    
	morecomment=[l]{;}, 
	alsoletter={:,-},   
	morekeywords={
		Event:,Observation:,PDDL-facts:,  	
		define,domain,problem,not,and,or,when,forall,exists,either,
		:domain,:requirements,:types,:objects,:constants,
		:predicates,:action,:parameters,:precondition,:effect,
		:fluents,:primary-effect,:side-effect,:init,:goal,
		:strips,:adl,:equality,:typing,:conditional-effects,
		:negative-preconditions,:disjunctive-preconditions,
		:existential-preconditions,:universal-preconditions,:quantified-preconditions,
		:functions,assign,increase,decrease,scale-up,scale-down,
		:metric,minimize,maximize,
		:durative-actions,:duration-inequalities,:continuous-effects,
		:durative-action,:duration,:condition
	}
}

\title{Planning from video game descriptions}

\usepackage{authblk}

\author[1]{Ignacio Vellido}
\author[1]{Carlos N{\'u}\~{n}ez-Molina}
\author[ ]{Vladislav Nikolov}
\author[1]{Juan Fdez-Olivares}

\affil[1]{Dpto. Ciencias de la Computación e I.A, University of Granada, Spain}

\affil[ ]{}

\affil[ ]{\textit{ignaciovellido@ugr.es, ccaarlos@correo.ugr.es, vladis890@gmail.com, \{faro, fgr\}@decsai.ugr.es}}

\renewcommand{\shorttitle}{Planning from video game descriptions}

\hypersetup{
pdftitle={Planning from video game descriptions},
pdfsubject={q-bio.NC, q-bio.QM},
pdfauthor={I. Vellido, C. N{\'u}\~{n}ez-Molina, V. Nikolov and J. Fdez-Olivares},
pdfkeywords={Knowledge Engineering, Automated Planning, Artificial General Intelligence},
}

\begin{document}
\maketitle

\begin{abstract}
This project proposes a methodology for the automatic generation of action models from video game dynamics descriptions, as well as its integration with a planning agent for the execution and monitoring of the plans. Planners use these action models to get the deliberative behaviour for an agent in many different video games and, combined with a reactive module, solve deterministic and no-deterministic levels. Experimental results validate the methodology and prove that the effort put by a knowledge engineer can be greatly reduced in the definition of such complex domains. Furthermore, benchmarks of the domains has been produced that can be of interest to the international planning community to evaluate planners in international planning competitions.
\end{abstract}


\section{Introduction}

Video games has long been a good playground for AI techniques as a way to collect data for real world problems. Over the years researchers has put a strong focus on AI applications in video games due to their easiness of modelling, visualisation and understanding while keeping common characteristics from real life situations.

Likewise, automated planning \citep{automatedPlanning} has proven to be a powerful branch when optimising sequences of action for intelligent agents. With only prior knowledge of the world dynamics and the actual state, planning techniques are capable of finding action plans to achieve the desired goal. Concerning video games, the use of automated planning to guide the behaviour of either an automated player or Non Playable Characters (NPCs) has been previously addressed \citep{kelly2008offline,hoang2005hierarchical} providing several advantages, among them that the agent is endowed with deliberative reasoning capable of solving problems in the video game, thus increasing the cognitive capabilities of automated players.

However, an important obstacle hinders the adoption of this technology: the creation of a planning domain that perfectly represents the world (in our case, a dynamics of a video game) is a long and difficult task. This knowledge engineering process requires to model in a planning language (like PDDL \citep{pddl1.2}, actual de facto standard) the world objects, their properties and relations, as well as the actions and how they transform the environment.

We believe that employing a language closer to the application field combined with a compiling process to a planning language the required effort to integrate planning techniques for problem solving can highly be reduced. With this methodology the knowledge engineer only needs to focus on the abstraction and codification of the information expressed in the starting language as planning knowledge, while experts can comfortably work on defining the requirements of the problems.

To display this idea, we are using a simple but expressive language such as VGDL (Video Game Description Language) \citep{vgdl1} to formally describe the objects and dynamics of the video game, as well as the different levels or scenarios. This language has been used to defined more than 100 video games in the GVGAI environment \citep{gvgai}, a framework for testing techniques oriented to Artificial General Intelligence (AGI). 

Our aim is to seamlessly integrate a classical planner into the GVGAI framework while reducing at the minimum the knowledge engineering effort to represent planning knowledge. This automated process enables planning techniques to be tested and compared with other AGI techniques that have already been successfully proved on previous competitions in this environment \citep{torrado2018deep}.

Therefore, our main contribution is triple: (1) we propose an automated knowledge based process that, receiving a VGDL game description, produces a PDDL domain that represents the game objects, their relationships and the dynamics; (2) this methodology is integrated in a planning based deliberative architecture to guide the behaviour of an agent in the games, combined with a reactive component to monitor the execution in non-deterministic games; (3) given the challenging nature of the domains we conducted tests on five planners following the format of the International Planning Competition 2018 that can be of interest to the international planning community.

This methodology opens the door to use any PDDL-based planner in an arbitrarily large number of video games, and whose behaviour can be compared with AGI techniques previously employed in GVGAI. Furthermore, due to their complex nature, the domains produced can be use as a testbed for future International Planning Competitions, as video game inspired domains are usually challenging and difficult to define.

The remainder of this article is structured as follows. In the next section we introduce some background concepts on GVGAI, VGDL, and PDDL. In Section \ref{related_work} we discuss previous attempts on using planning in video games, and Section \ref{methodology} formally describes the components of our proposed methodology. It is followed by a description of the validation process and the experimental results for multiple planners in Sections \ref{validation} and \ref{benchmarks}. Finally, this paper concludes in Section \ref{conclusions} with conclusions and future work.


\section{Background}\label{background}

\subsection{GVGAI and VGDL}

GVGAI (General Video Game AI) \citep{gvgai} is a framework for evaluating the performance and generalisation capabilities of AI-based techniques in multiple domains, including more than 100 video games and levels of all kind. Although the framework is oriented to address problems of General Artificial Intelligence, we aim to use this framework as a workbench for planning techniques, specially for the integration of planning and acting.

The underlying language that GVGAI uses for video games descriptions is VGDL (Video Game Description Language) \citep{vgdl1}, a high-level language focused on speed and simplicity. There is a large number of predefined types in VGDL, ranging from static ones to different classes of NPCs and agents, making possible the definition of many kinds of video games.

A game in VGDL is defined by two files, a Game Description File and a Level Description File, from now on respectively referred as GDF and LDF. A GDF specifies the dynamics, details the objects and establish relationships. A LDF indicates the layout of objects instances in a level.


\begin{figure}
\centering
\begin{lstlisting}[basicstyle=\small\ttfamily]
            SpriteSet
                player   >  FlakAvatar   stype=bullet
                alien    >  Bomber       stype=bomb
                missile  >  Missile
                    bullet  >  orientation=UP    
                    rock    >  orientation=DOWN
                
            LevelMapping
                p  <  player
                a  <  alien
                b  <  bullet
                r  <  rock
                
            InteractionSet
                player  rock   <  killIfFromAbove
                bullet  rock   <  killBoth
                alien  bullet  <  killSprite
                
            TerminationSet
                SpriteCounter  stype=player  limit=0  win=False
                SpriteCounter  stype=alien   limit=0  win=True
\end{lstlisting}
\caption{Example of a VGDL GDF. In this game, like in the Atari classic \textit{Space Invaders}, the player must shoot all the aliens without getting hit by a rock.}
\label{vgdl_example}
\end{figure}

\subsubsection{VGDL Game Description File}

The GDF from a VGDL game is composed of four different sections. Following the example of Figure \ref{vgdl_example}, we proceed to explain their meaning:

\begin{itemize}    
    \item \textbf{SpriteSet}: A specification of the objects involved in the game (avatar, enemies, missiles, etc., sometimes referred as \textit{sprites}). This representation can be hierarchical, allowing the inheritance of attributes and behaviour. The definition of an object is formed by a name, a predefined type, and a list of arguments that precises the behaviour of the object.
    
    In our example we include four objects, two inheriting from a common parent. From top to bottom, in the first place we define the agent as a \textit{FlakAvatar}, an avatar capable of horizontal movements and shooting. Then, enemies are defined as \textit{Bomber}, \textit{i.e.} NPCs who randomly shoot an object that, as well as with the agent, is specified in their parameters. Lastly, the final two objects correspond to the missiles shot by the avatar and the enemies, each one with a different orientation.
    
    \item \textbf{LevelMapping}: Characters and the objects they stand for in a LDF. For example, in the figure we can see that \textit{player} and \textit{alien} instances are respectively represented as \textit{p} and \textit{a}.

    \item \textbf{InteractionSet}: A specification of the interactions between the objects, indicating which actions the game engine needs to execute when they collide with each other.
    As with the different kinds of sprites, there is a previously defined repertory of interactions in VGDL. The order in which the objects are specified is relevant, as the latter is responsible of the interaction and the former receives the main effects.
    
    In our example we have three different interactions. The first one models the death of the agent when a rock falls from above, the second one says both bullets and rocks get destroyed upon collision, and the last one indicates that an alien gets killed if it is hit by a bullet.

    \item \textbf{TerminationSet}: The list of criteria that make the game finalise. In our case we have both a condition for a victory and a lost, depending on whether the player dies or not before the aliens do.
\end{itemize}


\begin{figure}[t]
	\centering
	\includegraphics[width=0.6\textwidth]{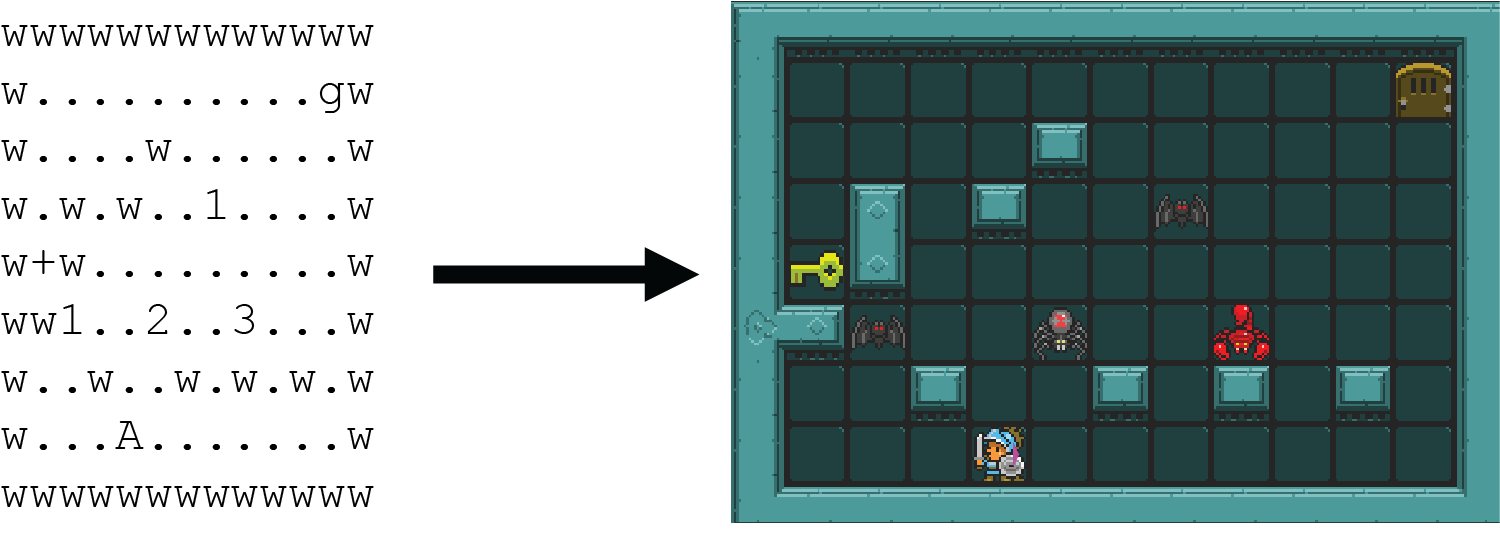}
    \caption{Level representation in GVGAI. Notice how each character represents an instance of an object. For example: ``\textit{w}'' means \textit{wall}, ``.'' means a free cell, ``\textit{A}'' means \textit{avatar} and each number a different enemy.}
    \label{level}
\end{figure}

\subsubsection{VGDL Level Description File}

A level description file is composed of a 2D matrix of characters where each cell indicates the starting position of an object instance. Each character is associated with its object as stated in the \textit{LevelMapping} section. Figure \ref{level} shows a transformation from the level description to the visual representation in GVGAI.


\subsection{PDDL}

Automated planning is a branch of A.I. concerned with the study of agent acting techniques. Two elements are required in a planning environment:
\begin{itemize}
    \item The action model available for the agent and the dynamics of the world (\textit{i.e.} how the actions of the agent modify its environment), referred as \textbf{domain}.
    \item A description of the initial state of the world, the objects involved in it and the desired goals, called \textbf{problem}.
\end{itemize}

Both planning domains and problems are formally described using first order logic, and over the years PDDL (Planning Domain Description Language) \citep{pddl1.2} has been accepted as a standard language. A PDDL domain must contain the following sections:
\begin{itemize}
    \item \textbf{Types}: Hierarchical representation of the kind of objects present in the world.
    \item \textbf{Predicates}: Relations or characteristics between objects.
    \item \textbf{Actions}: Transformations in the state of the world, formed by three sections: (i) a series of \textbf{parameters}, \textit{i.e.} objects involved in the action; (ii) a set of \textbf{preconditions}, \textit{i.e.} predicates needed to be true before performing the action; (iii) the list of \textbf{effects}, describing the changes produced in the world after the action is completed.
\end{itemize}

PDDL has been supported and extended with new versions and functionality, like numeric predicates and continuous and durative actions \citep{pddl2.1}. Although for our purposes some of this features can simplify the knowledge representation, in order not to reduce compatibility with older planners we stayed within the original 1.2 properties.


\section{Related work}\label{related_work}

Regarding video games, automated planning has long been considered an enabling technology to control the behaviour of deliberative agents. \citep{vcerny2016plan} shows an in-depth review of related work about planning and video games. Nevertheless, up to the authors knowledge, most works are focused on evaluating the performance of planners, and planning domains are manually represented. In most cases, only a single planning domain is represented and planner performance is evaluated on different problems (or game scenarios/levels). On the other hand, approaches like \citep{geffner2015width} go further, and show that planning algorithms are competitive against standard techniques in Atari video games, but without a compact PDDL-model for those games.

For video game frameworks, several planning architectures have been integrated in StarCraft \citep{martinez2018goal} or Minecraft \citep{bonanno2016selecting}, based on the idea of evaluating the performance of an online planning architecture using a single planning domain. Concerning GVGAI, some approaches provide results that allow reactive controllers to learn with relative success \citep{torrado2018deep}. However, these controllers lack  deliberative capabilities and employ black box models, making impossible the understanding of the reasoning behind the agent actions.

As previously mentioned, we are concerned with the video game framework GVGAI. To the best of our knowledge, the only attempt of applying planning to this framework is \citep{couto}, where authors manually define a controller for puzzle games using PDDL. The approach here presented tries to go one step further, by proposing a methodology to automatically generate planning domains from mostly any video game described in VGDL in this framework, integrated with a planning-agent in order to control the behaviour of an automated player.

\vspace{\baselineskip}

We have previously faced the generation of HTN models from video game descriptions \citep{ignacio} but, as opposed to classical planning approaches, HTNs need the inclusion of a learnt agent strategy to solve the games. To surpass these additional effort, in this project we employ PDDL as a method to let the planner decide the best course of action for the agent.

Lastly, generation of planning domains from descriptions represented in domain-specific languages has been addressed previously in approaches like \citep{eLearning} for eLearning scheduling, \citep{BPM} for Business Process Management, and \citep{clinical} for supporting clinical processes. All these works use a language related to the application field and generate planning domains that support expert decision making. The novelty in our methodology against these approaches is the use of abstract templates that encode the knowledge expressed in the domain language as planning knowledge, opening the door to define as many cases of study as the domain-specific language is capable of express.


\begin{figure}[t]
	\centering
    \includegraphics[width=\textwidth]{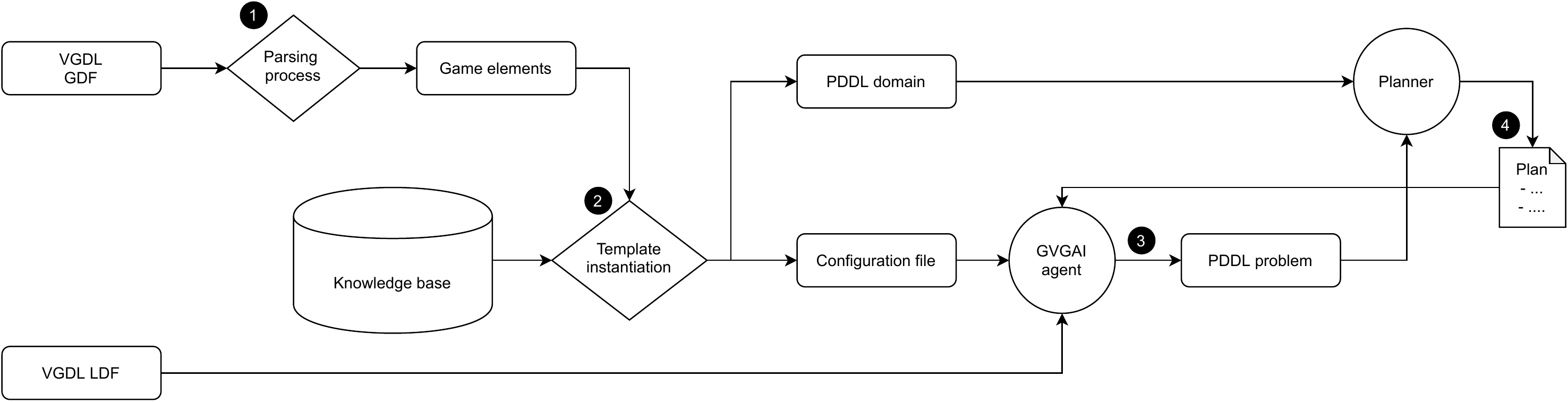}
    \caption{General view of our approach.}
    \label{process}
\end{figure}

\section{Methodology}\label{methodology}

The automated generation process starts from the GDF and LDF files describing, respectively, a VGDL video game and a specific level of that game. As a result, a PDDL domain is produced and served as input for a planning agent that, until the objectives are achieved, repeatedly plans and acts. 

The domains produced are deterministic, because modelling uncertainly will increase a complexity that it is by itself already high. Instead, to cope with no-deterministic situations a reactive module will search for inconsistencies between the planning and game state, replanning when needed.

The methodology, represented in Figure \ref{process}, is divided of four main steps, further described in the following subsections.


\subsection{Extraction of game elements}

As the first step in the methodology the GDF is parsed\footnote{Using a parsing process based on the ANTLR \citep{antlr} parsing facilities.} and a set of game elements are extracted from each section. Each element encode an individual piece of a GDF, independently of the rest of components of the game, and will be later conveniently translated into elements of the PDDL domain.

Different information is stored depending on which GDF section they come from:

\begin{itemize}    
    \item \textbf{SpriteSet}: Name and type of each object are parsed from this section. Additionally, because the objects are structured hierarchically in the GDF, it is necessary to maintain this hierarchy between elements in the PDDL domain. Lastly, functional parameters (like speed or movement directions, not the ones concerning the visual representation) are also stored to specify the properties of the objects.

    \item \textbf{LevelMapping}: The correspondences between the characters in the level description and the specific instances of the game objects are extracted in order to be able to produce the PDDL problem. This codification will be passed to the agent via a configuration file to associate each character in a game observation (\textit{i.e.}, the LDF of the game at any turn) to the corresponding predicates and types.
    
    \item \textbf{InteractionSet}: Type and objects involved in each interaction are parsed, as well as the role of each object of the pair, either the \textit{producer} or the \textit{receiver} of the action represented in the interaction.
    
    \item \textbf{TerminationSet}: Concerning ending criteria we store the objects involved and their number, as well as whether the agent wins or lose under that condition.
\end{itemize}


\subsection{Domain generation}\label{compiler}

Once all the game elements are known, a VGDL to PDDL compiling process produce the PDDL domain as well as a configuration file. This additional file encodes the knowledge required to produce the PDDL problem from the initial game state and to monitor the execution of a plan, and is further described in the next subsection.

The domain is formed by the instantiation of PDDL templates stored in a knowledge base. According to the parsed types of each game element, this module outputs and joins the different sections of a PDDL domain (types, predicates, and actions).


\begin{table}
    \centering
    \begin{tabular}{ll|ll}
    \multicolumn{2}{c}{\textbf{Game element}} & \multicolumn{2}{|c}{\textbf{Knowledge base}} 
    \\ \hline
    Name:              & $<$T$>$               & Actions:    & \begin{tabular}[c]{@{}l@{}}move\_$<$O$>$(...)\\ move\_stop(...)\end{tabular} \\
    
    Type:              & Missile              &            & \\
    Parameters:        & Orientation: $<$O$>$    & Predicates: &
    
    \begin{tabular}[c]{@{}l@{}}(missile\_$<$T$>$\_moved ?object)\\ (turn\_$<$T$>$\_move)\\ (finished\_turn\_$<$T$>$\_move)\end{tabular}
    \end{tabular}

    \caption{Relation between a \textit{missile} type object and the information stored in the knowledge base. When the $<$T$>$ name and $<$O$>$ orientation are discovered in the parsing process the templates will be instantiated and included in the PDDL domain.}
    \label{game_element_and_templates}
\end{table}

\begin{table}[t]
    \centering
    \begin{tabular}{c|c}
\begin{lstlisting}[
            %float=!tb,
            language=PDDL]
(:action <@\textcolor{red}{<S1>}@>_<@\textcolor{blue}{<S2>}@>_COLLECTRESOURCE
  :parameters(?o1 - <@\textcolor{red}{<S1>}@> ?o2 - <@\textcolor{blue}{<S2>}@>
             ?x ?y ?r ?r_next - num)
  :precondition (and
      (turn-interactions)

      ; Verify objects are different
      (not (= ?o1 ?o2))
      (at ?x ?y ?o1)
      (at ?x ?y ?o2)

      (got-resource-<@\textcolor{red}{<S1>}@> ?r)
      (next ?r ?r_next)
  )
  :effect (and
      ; Remove resource from map
      (not (at ?x ?y ?o1))
      (dead ?o1)

      ; Increase value
      (not (got-resource-<@\textcolor{red}{<S1>}@> ?r))
      (got-resource-<@\textcolor{red}{<S1>}@> ?r_next)
  )
)
\end{lstlisting} &
\begin{lstlisting}[
            %float=!tb,
            language=PDDL]
(:action <@\textcolor{red}{SHOES}@>_<@\textcolor{blue}{USER}@>_COLLECTRESOURCE
  :parameters(?o1 - <@\textcolor{red}{shoes}@> ?o2 - <@\textcolor{blue}{user}@>
             ?x ?y ?r ?r_next - num)
  :precondition (and
    (turn-interactions)

    ; Verify objects are different
    (not (= ?o1 ?o2))
    (at ?x ?y ?o1)
    (at ?x ?y ?o2)

    (got-resource-<@\textcolor{red}{shoes}@> ?r)
    (next ?r ?r_next)
  )
  :effect (and
    ; Remove resource from map
    (not (at ?x ?y ?o1))
    (dead ?o1)

    ; Increase value
    (not (got-resource-<@\textcolor{red}{shoes}@> ?r))
    (got-resource-<@\textcolor{red}{shoes}@> ?r_next)
  )
)
\end{lstlisting}
\end{tabular}

\caption{Interaction depicting how a player can increase its pool of resources, On the left we show the template stored in the knowledge base. When the interaction is detected and the objects involved known, the compiler insert the appropriate names, producing the PDDL action shown on the right.}
\label{interaction_example}
\end{table}

\subsubsection{Knowledge base}

The knowledge base contains templates to produce all the parts that forms a PDDL domain, as well as the initial predicates that need to be included in a PDDL problem. In order to achieve as much as generalisation is possible, these templates are game-independent as they are only related to the specific concept they represent.

Generally speaking, there are three kinds of templates:

\begin{itemize}    
    \item \textbf{Templates for sprites}: Concerned with the representation of objects in the game (excluding the avatar). Each sprite results in a different PDDL \textit{type} and, depending on its VGDL type, additional actions and predicates may be included to correctly model its behaviour.
    
    As we can see in Table \ref{game_element_and_templates}, a \textit{missile} type object will produce predicates and actions to update its movement in each game turn, moving only in the direction indicated in its parameters. 
    
    \item \textbf{Templates for avatars}: This section is concerned with the subset of sprites that represents the automated players in the game. The actions represented for avatars in the knowledge base are aimed to tile-based games with grid physics, including moves in the four cardinal directions and the possibility of using a previously defined resource (for example, a sword to kill enemies).
        
    The knowledge base stores templates for each possible action of a VGDL avatar and knows which ones are available for each type.
    
    \item \textbf{Templates for interactions}: The last section stores templates for VGDL interactions, (\textit{i.e.} the effect of a collision between two objects). Each one has particular PDDL action templates, with the resulting instantiation as shown in Table \ref{interaction_example}.
\end{itemize}

The process of defining a template consist of representing the knowledge of how the intended concept functions (in our case, and action or an interaction) independently of the rest of the elements in the game. When the intention is to define only a domain, repeating this process for each element in the language is not desirable, as the time consumed is greater than to directly model the specific domain. However, when it is expected to work with multiple related domains the entire effort is considerably reduced. Since each template is abstract, once it has been defined and validated it can be reused as many times as the user wants without intervention of the knowledge engineer.


\begin{table}[t]
    \centering
    \begin{tabular}{c|c}
\begin{lstlisting}[
            %float=!tb,
            language=PDDL]
(:action BOULDER_MOVE_DOWN
  :parameters (?o - boulder 
              ?x ?y ?new_y - num)
  :precondition (and
    (turn-boulder-move)
    (not (boulder-moved ?o))
    (oriented-down ?o)
          
    (at ?x ?y ?o)
    (next ?y ?new_y)
          
    ; Check cell is empty
    (not (is-wall ?x ?new_y))
  )
  :effect (and
    (not (at ?x ?y ?o))
    (at ?x ?new_y ?o)
    
    (boulder-moved ?o)
  )
)
...
\end{lstlisting} &
\begin{lstlisting}[
            %float=!tb,
            language=PDDL]
(:action STOP_BOULDER_MOVE
  :parameters ()
  :precondition (and
    (turn-boulder-move)
    
*   (forall (?o - boulder) 
*     (or (dead ?o) 
*         (boulder-moved ?o))
*   )
  )
  :effect (and
    (forall (?o - boulder)
      (not (boulder-moved ?o))
    )

    (not (turn-boulder-move))
    (finished-turn-boulder-move)
  )
)
\end{lstlisting}
\end{tabular}

\caption{PDDL actions that model the behaviour of a \textit{missile} type object. The first action is called repeatedly checking the orientation of an instance and setting their position accordingly. Once all missiles have moved, the marked predicate becomes true and the planner continues with the rest of the turn.}
\label{missile_action}
\end{table}

\subsubsection{Constructing the domains}

As previously mentioned, the templates returned from the knowledge base are instantiated with the help of the game elements, obtaining as a result a domain formed by the combination of all the templates. Each PDDL section in the domain is constructed differently:

\begin{itemize}
  \item \textbf{Types}: For each child sprite, the hierarchy is reflected defining the parent as the object type. When reached a sprite without parent, its type defined in the GDF will be its PDDL type. All sprite types inherit from the most generic object, a supertype called \textit{Object}, allowing us to generalise behaviour in predicates and actions.
  
  To reduce the complexity of a domain (in the number of interactions and possible sprites instances), static objects are instead modelled as a predicate \textit{(is-$<$T$>$ ?x ?y)}. Although this method is not a faithful representation of VGDL, it does not harm the generalisation capabilities of the methodology. On the contrary, the additional knowledge optimises the domains with significant results\footnote{This kind of optimisation has shown us a reduction in the number of instances in a PDDL problem up to a 75\%.} making possible to plan and act in bigger and complex levels.

  \item \textbf{Predicates}: We keep multiple kinds of predicates:
  \begin{itemize}
      \item The actual position of each instance.
      \item The orientation of every instance capable of movement (avatars and non-static sprites).
      \item Specific predicates for resources, like coins, gems, or boots.
      \item To force the execution order between actions (further detailed in Subsection \ref{turn_section}).
      \item And finally, because PDDL 1.2 lacks numerical values for resources and coordinates, we simulated them with a new type \textit{num} and predicates of order (\textit{e.g. (next ?n1 ?n2 - num)}). That way, connectivity within cells is implicitly expressed and only positional predicates are needed.
  \end{itemize}

  \item \textbf{Actions}: The generation of actions is structured as follows:
  \begin{itemize}
      \item For each possible movement of the avatar an action is generated, plus one for the no-movement. The latter is included in case the agent needs for an event to occur before making its move.
      \item For non-static sprites we add actions that change their state each turn, as previously shown in Table \ref{missile_action}.
      \item For each interaction an action reproduce its effects, as in Table \ref{interaction_example}. 
      \item Additional actions that model the turn order of a GVGAI game are defined. Primarily, these actions make sure all non-static objects are updated and all interactions are resolve before beginning the following turn.
  \end{itemize}
\end{itemize}


\begin{figure}[t]
    \centering
    \includegraphics[width=0.7\textwidth]{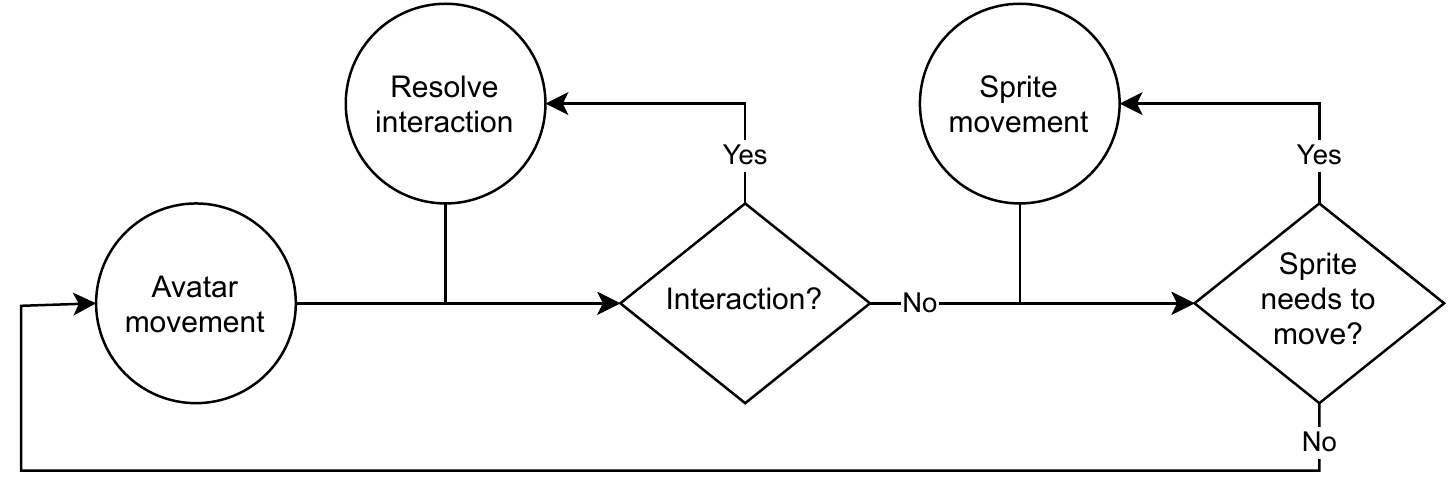}
    \caption{Actions that model the game turn in the resulting domains.}
    \label{turn}
\end{figure}

\begin{table}
  \begin{framed}
      \centering
      \begin{lstlisting}[
          %float=!tb, 
          language=PDDL]
Sokoban: Pushing a box into a hole

 + AVATAR_ACTION_MOVE_DOWN            (avatar n3 n4 n5)
 - BOX_AVATAR_BOUNCEFORWARD_DOWN      (box_3_4 avatar n3 n5 n6)
 - BOX_HOLE_KILLSPRITE                (box_3_4 hole_3_6 n3 n6)
 - END-TURN-INTERACTIONS              ()
 # END-TURN-SPRITES                   ()
--------------------------------------------------------------------------
IceAndFire: Grabing a resource

 + AVATAR_ACTION_MOVE_RIGHT           (avatar n12 n1 n13)
 - ICESHOES_AVATAR_COLLECTRESOURCE    (iceshoes_13_1 avatar n13 n1 n0 n1)
 - END-TURN-INTERACTIONS              ()
 # END-TURN-SPRITES                   ()
--------------------------------------------------------------------------
Boulderdash: Digging underneath a boulder

 + AVATAR_ACTION_USE_RIGHT            (avatar n11 n6 n12)
 - DIRT_SWORD_KILLSPRITE              (dirt17 sword1 n12 n6)
 - END-TURN-INTERACTIONS              ()
 # SWORD_DISAPPEAR                    (sword1 n12 n6)
 # STOP_SWORD_DISAPPEAR               ()
 # BOULDER_MOVE_DOWN                  (boulder3 n12 n5 n6)
 # STOP_BOULDER_MOVE                  ()
 # END-TURN-SPRITES                   ()
      \end{lstlisting}
  \end{framed}
  \caption{Output plans for one turn in different games. (+) indicates avatar movements, (-) interactions and (\#) other sprite movements.}
  \label{plans}
\end{table}

\subsubsection{Modelling the game turn}\label{turn_section}

Each game running in the GVGAI environment follows a fixed sequence of steps each turn, and because our domains are intended to work in this framework they must do the same. Due to the lack of explicit mechanisms to order the evaluation of actions in PDDL, we need to define an alternate set of actions and predicates to ensure the order is followed every turn.

The steps of a turn in our domains are the following, as shown in Figure \ref{turn} and exemplified in Table \ref{plans}:

\begin{enumerate}
    \item Firstly, an action for the agent is chosen. The planner and its heuristic are responsible of selecting between the available moves the one best suited to accomplish the goal.
    
    \item Secondly, all interactions are resolved before continuing with the turn execution. Meaning that no pair of object with an interaction defined in the GDF are actually colliding in the game state.
    
    \item Similarly, a set of control actions force each non-static object to perform a movement and update its state. For example, the marked preconditions in the action on the right of Table \ref{missile_action} will only become true when each \textit{missile} type instance has already moved.
\end{enumerate}


\begin{table}[t]
    \centering
    \begin{tabular}{p{2.4cm}|l|l}
\textbf{Sokoban}:\newline
Put every box in a hole &
\begin{lstlisting}[
            %float=!tb,
            language=PDDL]

InteractionSet
  box hole > killSprite
            
TerminationSet
  SpriteCounter stype=box limit=0  
                win=True
  
\end{lstlisting} &
\begin{lstlisting}[
            %float=!tb,
            language=PDDL]
(:goal
  (forall 
    (?o - box) 
    (dead ?o)
  )
)
\end{lstlisting} \\ \hline

\textbf{Boulderdash}:\newline
Get 9 gems and go to the exit &
\begin{lstlisting}[
            %float=!tb,
            language=PDDL]
            
InteractionSet
  exit avatar > killIfOtherHasMore 
                resource=gem limit=9
            
TerminationSet
  SpriteCounter stype=exit limit=0  
                win=True

\end{lstlisting}  & 
\begin{lstlisting}[
            %float=!tb,
            language=PDDL]
(:goal
  (forall 
    (?o - exit) 
    (dead ?o)
  )
)
\end{lstlisting} \\
\hline
\textbf{Waves}:\newline
Survive during 50 turns &
\begin{lstlisting}[
            %float=!tb,
            language=PDDL]
            
TerminationSet
  Timeout limit=50 win=True

\end{lstlisting}  &
\begin{lstlisting}[
            %float=!tb,
            language=PDDL]
(:goal
  (and
    (turn n50)
    (not 
      (dead avatar)
    )
  )
)
\end{lstlisting} \\
\end{tabular}
\caption{Ending criteria for three games and their VGDL and PDDL representations.}
\label{goals}
\end{table}

\subsubsection{Goal deduction}

VGDL games are goal oriented (as opposed to reward oriented), making them suitable for classical planning representation. Although there are various types of \textit{TerminationCriteria}, most of them are variations of \textit{SpriteCounter}. This condition gets activated when the number of instances of an object reach a specific threshold.

Despite the fact that using one kind of criteria may seem limited, with the right combination of interactions it does not restrict the expressiveness of the language when defining objectives. For example, in Table \ref{goals} we compare different game goals with their VGDL representation.

As for our PDDL representation, all this goals can be modelled with a \textit{forall} condition checking no instance is present in the level (but with the inclusion of numeric predicates a simple counter will be preferred).

Even though technically there is another kind of ending criteria, there are few games who actually use it. This type of goal, called \textit{Timeout}, activates when the game reach the turn indicated in the parameters. Nevertheless, it can be easily codified in PDDL adding a new counter predicate who checks how many turns have passed since the beginning of the execution.


\begin{table}[t]
    \centering
    \begin{tabular}{|c|c|c|}
\hline
\textbf{Configuration file} & \textbf{LDF} & \textbf{PDDL problem} \\
\hline
\begin{lstlisting}[
            %float=!tb,
            language=PDDL]
gameElementsCorrespondence:
  avatar:
  - (at ?x ?y ?avatar)
  hole:
  - (at ?x ?y ?hole)
  box:
  - (at ?x ?y ?box)
  wall:
  - (is-wall ?x ?y)
variablesTypes:
  ?hole: hole
  ?avatar: avatar
  ?box: box
  ?x: num
  ?y: num
goals:
  - goalPredicate: 
      (forall (?o - box) 
      (object-dead ?o)
      )
    priority: 1
\end{lstlisting} &
\begin{lstlisting}[
            %float=!tb,
            language=PDDL,
            basicstyle=\fontsize{15}{13}\selectfont\ttfamily]
wwwww
wh  w
w b w
w A w
wwwww
\end{lstlisting} &
\begin{lstlisting}[
            %float=!tb,
            language=PDDL]
(define (problem SokobanProblem)
  (:domain SokobanDomain)
  (:objects
    avatar - avatar
    box_2_2 - box
    hole_1_1 - hole
    n0 n1 n2 n3 n4 - num
  )
  (:init
    (at n2 n2 box_2_2)
    (at n2 n3 avatar)
    (at n1 n1 hole_1_1)
    (is-wall n0 n0)
    (is-wall n1 n0)
    ...
  )
  (:goal
    (forall (?o - box)
      (object-dead ?o)
    )
  )
)
\end{lstlisting} \\ \hline
\end{tabular}
\caption{Example of a configuration file, a LDF, and the consequent translation into a PDDL problem for Sokoban, a game where the agent must push all the boxes into holes.}
\label{configuration_file}
\end{table}

\subsection{Problem generation}\label{problem_generation}

As we are looking to define a continuous process of replanning, we let the agent be responsible of the PDDL problem definition. In order to do that, the VGDL compiler generates and additional file, called configuration file, that encodes the knowledge required to produce a PDDL problem from a LDF.

In Table \ref{configuration_file} we illustrate the content of a configuration file and the resulting problem from a LDF. This file is composed of three segments:
\begin{itemize}
    \item \textbf{gameElementsCorrespondence}: Indicates which predicates should be included in the problem when an instance of that specific GVGAI type is detected.
    
    \item \textbf{variablesTypes}: Saying to what PDDL types correspond the variables appearing in the segment above. Each instance will get represented with a distinct PDDL object in the problem.
    
    \item \textbf{goals}: Objectives for the game, who will be copied as they are defined into the problem.
\end{itemize}


\subsection{Planning and acting}

The methodology exposed above shows the automatic generation of PDDL domains and problems from a video game description. These files can be used by itself as a workbench for planners due to their complex nature. However, if the intention is to develop a planning agent to solve video games, there are some further points needed of consideration.

\subsubsection{Goal scheduling}

Although the parsed goals from the GDF only model the ending criteria for the game as a unique PDDL objective, we consider that some kind of subgoal selection could be a promising inclusion on this methodology. This will not only help the agent to face non-deterministic situations, but will also make each planning step faster and plans shorter. We suggest some possible ideas in the last section of this paper.


\subsubsection{Monitoring execution and replanning}

Since the domains produced are deterministic, occasionally in non-deterministic games plans cannot be carried out. Maybe a rock has dropped in an unexpected cell, or enemies have appeared in a path that if the agent tries to follow it would not survive. 

For those cases, we included a monitoring module that checks before sending each avatar action to the GVGAI framework that all the preconditions in the PDDL action are still consistent with the actual game state. If at least one of them is not, the actual plan is discarded and a new PDDL problem is generated following the same schema as described in section \ref{problem_generation}, using the last game state as a LDF. Afterwards, the original PDDL domain and the new problem are sent to the planner, who finds a new plan for the agent. The change in the veracity of the preconditions ensures that the same movement will not be chosen again.


\section{Knowledge validation}\label{validation}

The validation of a planning domain is a knowledge validation task, but while a knowledge based system is frequently intended to classification or diagnosis, in planning is aimed to build plans. Thus, the knowledge in our model represent actions and how these affect the objects in the world \citep{mccluskey}.

Due to the lack of formality in the requirements of our system, as they are the VGDL language and the semantic meaning of its elements, who can not be mathematically expressed, we decided to carry out a validation by inspection process with unit tests, a extended methodology from software validation consisting in evaluate individual units of source code whose expected behaviour is known.

Therefore, we carried out multiple planning test cases (unit tests) over each template of our knowledge base, each one formed by a domain with the specific template to verify (\textit{e.g.} an action modelling an interaction, a movement of a missile, an avatar, etc.) and the smallest problem possible to carry out the test. The verification process consisted on ascertain that at every step the game state was the expected in the planning state.

Furthermore, thanks to the graphical interface provided in GVGAI we carried out an additional visual validation of the agent execution, running the planning agent multiple times over different levels following the game execution in the planner state and verifying that the game dynamics were correctly represented and the objectives solved without errors.

In the video \url{https://youtu.be/F7kg6LC-sDI} we show the agent successfully solve level instances in four games: Zenpuzzle, Sokoban, IceAndFire and Butterflies.


\begin{table}[t]
  \centering
  \resizebox{.9\columnwidth}{!}{%
  \makebox[\textwidth][c]{
  \begin{tabular}{| c | c c c c | c c c |} 
      \hline
      & \multicolumn{4}{c|}{Syntactic elements} & \multicolumn{3}{|c|}{Semantics elements} \\
      & Types & Supertypes & Predicates & Actions  & Determinism & Agent actions & Interactions \\
      \hline
      Bait & 7 & 4 & 15 & 15 & Yes & 4 & 8 \\
      \hline
      Boulderdash & 11 & 8 & 21 & 23 & No & 12 & 8 \\
      \hline
      Butterflies & 4 & 3 & 13 & 8 & No & 4 & 1 \\
      \hline
      CakyBaky & 11 & 4 & 22 & 17 & No & 8 & 6 \\
      \hline
      ChipsChallenge & 19 & 4 & 27 & 26 & Yes & 4 & 19 \\
      \hline
      IceAndFire         & 8 & 4 & 18 & 10 & Yes & 4 & 3 \\
      \hline
      SimplifiedBoulder. & 7 & 6 & 17 & 20 & Yes & 12 & 5 \\
      \hline
      Sokoban            & 4 & 3 & 13 & 12 & Yes & 4 & 5 \\
      \hline
      Zelda              & 9 & 7 & 17 & 20 & No & 12 & 5 \\
      \hline
      Zenpuzzle          & 5 & 2 & 15 & 8 & Yes & 4 & 1 \\
      \hline
  \end{tabular}
  }}
  \caption{Syntactic and semantic elements of each game.}
  \label{domains}
\end{table}

\section{Experimental results}\label{benchmarks}

The main goal in our experimentation is to propose, following the format of the IPC 2018, multiple benchmarks of our domains, highlighting the behaviour of state-of-the-art planners in complex situations. The experiments were carried out on an Intel i7-8750H running at 2.20Ghz and 6GB of RAM, within a time limit of 15 minutes.

To that end, we chose 10 games from the GVGAI repertory, three puzzle and seven reactive games, containing different levels of determinism, non-determinism and characteristics as shown in Table \ref{domains}. Each game was evaluated in 10 levels, five predefined in GVGAI and an additional five we handcrafted, varying in map dimensions and number of instances.

The planners selection was based on the easiness of compilation in modern OS, their behaviour towards grounding, and the variety of techniques employed. The chosen planners were:
\begin{itemize}
    \item Madagascar \citep{madagascar}, applying SAT based techniques.
    \item Fast Downward \citep{fast_downward}, with LAMA and A* configurations. LAMA \citep{lama} is a multi-heuristic system with a weighted A* search.
    \item Saarplan \citep{saarplan}, based on decoupled search, an algorithm that reduces the search-space representation size by means of partitioning the state variables into components, forming a star topology and only searching in the centre component of the topology.
    \item LAPKT-DUAL-BFWS \citep{dual}, combining best-first width search with pruning.
\end{itemize}

As for the tracks, we used the same metrics applied in the International Planning Competition 2018, these are: 
\begin{itemize}
    \item \textbf{Coverage}: Measured as the total number of levels solved.
    
    \item \textbf{Satisficing}: Valuing the cheapest plans (\textit{i.e.} with the minimum number of steps). The score of a planner on a level is calculated as:
    \begin{equation}
    S = \frac{C}{C*}    
    \end{equation}
    
    where C and C* denote the length of the found plan and the reference plan. Since the non-deterministic characteristic makes impossible to calculate a ground truth plan, we cannot find the shortest plan of a level for some games. In those cases, the cheapest discovered plan found by any of the planners was used as reference.
    
    \item \textbf{Agile}: Each instance is measured following the formula:
    \begin{equation}
    A = \begin{cases}
         1 & T \leq 1 \\
         1 - \frac{log(T)}{log(900)} & 1 < T \leq 900 \\
         0 & T > 900 \\
        \end{cases}
    \end{equation}
    being $T$ the number of seconds employed to find a plan, and 900 indicating the time limit imposed upon the planners. The final score is the sum of the agile score across all levels.
\end{itemize}


\begin{table}
\centering
\begin{tabular}{lccccccccccc}
\textbf{Coverage} & \rotatebox{90}{\textbf{bait}} & \rotatebox{90}{\textbf{boulderdash}} & \rotatebox{90}{\textbf{butterflies}} & \rotatebox{90}{\textbf{cakybaky}} & \rotatebox{90}{\textbf{chipschallenge}} & \rotatebox{90}{\textbf{iceandfire}} & \rotatebox{90}{\textbf{simplified-bould.}} & \rotatebox{90}{\textbf{sokoban}} & \rotatebox{90}{\textbf{zelda}} & \rotatebox{90}{\textbf{zenpuzzle}}  & \rotatebox{90}{\textbf{SUM}} \\ \hline
DUAL-BFWS         & \textbf{10}        & \textbf{10}               & \textbf{10}          & \textbf{10}       & \textbf{9}     & \textbf{10}         & \textbf{10}& \textbf{10}      & \textbf{10}    & \textbf{10} & \multicolumn{1}{|c}{99}\\ \hline
Saarplan          & 0  & 0         & 10   & 10& 0              & 10  & 0          & 0& 10             & 1           & \multicolumn{1}{|c}{41}\\ \hline
FD (A*)           & 0  & 0         & 10   & 10& 0              & 10  & 0          & 0& 10             & 1           & \multicolumn{1}{|c}{41}\\ \hline
FD (LAMA)           & 0  & 0         & 8   & 10& 0              & 10  & 0          & 0& 10             & 0           & \multicolumn{1}{|c}{38}\\ \hline
Madagascar        & 1  & 0         & 8    & 0 & 0              & 8   & 0          & 1& 6              & 8           & \multicolumn{1}{|c}{32}\\ \hline
\end{tabular}
\caption{Results of coverage score.}
\label{coverage}
\end{table}
\begin{table}
\centering
\makebox[\textwidth][c]{
\begin{tabular}{lccccccccccc}
\textbf{Sat score} & \rotatebox{90}{\textbf{bait}} & \rotatebox{90}{\textbf{boulderdash}} & \rotatebox{90}{\textbf{butterflies}} & \rotatebox{90}{\textbf{cakybaky}} & \rotatebox{90}{\textbf{chipschallenge}} & \rotatebox{90}{\textbf{iceandfire}} & \rotatebox{90}{\textbf{simplified-bould.}} & \rotatebox{90}{\textbf{sokoban}} & \rotatebox{90}{\textbf{zelda}} & \rotatebox{90}{\textbf{zenpuzzle}}  & \rotatebox{90}{\textbf{SUM}} \\ \hline
DUAL-BFWS & \textbf{10.00}	& \textbf{10.00}	& 7.09	& 8.65	& \textbf{9.00}	& \textbf{10.00}	& \textbf{10.00}	& \textbf{10.00}	& 8.34	& \textbf{9.18}	& \multicolumn{1}{|c}{92.3}\\ \hline

FD (A*) & 0.00	& 0.00	& \textbf{10.00}	& \textbf{9.99}	& 0.00	& 10.00	& 0.00	& 0.00	& \textbf{9.96}	& 1.00	& \multicolumn{1}{|c}{41.0}\\ \hline

FD (LAMA) & 0.00 & 0.00 & 8.00 & 9.99 & 0.00 & 10.00 & 0.00 & 0.00 & 9.96 & 0.00 & \multicolumn{1}{|c}{38.0}\\ \hline

Saarplan & 0.00	& 0.00	& 7.84	& 8.08	& 0.00	& 9.83	& 0.00	& 0.00	& 9.59	& 0.94	& \multicolumn{1}{|c}{36.3}\\ \hline

Madagascar & 0.32	& 0.00	& 7.16	& 0.00	& 0.00	& 7.36	& 0.00	& 0.76	& 5.58	& 7.68	& \multicolumn{1}{|c}{28.9}\\ \hline
\end{tabular}}
\caption{Results for satisficing score.}
\label{satisficing}
\end{table}
\begin{table}
\centering
\begin{tabular}{lccccccccccc}
\textbf{Agile score} & \rotatebox{90}{\textbf{bait}} & \rotatebox{90}{\textbf{boulderdash}} & \rotatebox{90}{\textbf{butterflies}} & \rotatebox{90}{\textbf{cakybaky}} & \rotatebox{90}{\textbf{chipschallenge}} & \rotatebox{90}{\textbf{iceandfire}} & \rotatebox{90}{\textbf{simplified-bould.}} & \rotatebox{90}{\textbf{sokoban}} & \rotatebox{90}{\textbf{zelda}} & \rotatebox{90}{\textbf{zenpuzzle}}  & \rotatebox{90}{\textbf{SUM}} \\ \hline
DUAL-BFWS & \textbf{5.97}	& \textbf{5.75}	& 8.17	& 5.59	& \textbf{5.94}	& \textbf{10.00}	& \textbf{6.62}	& \textbf{5.43}	& 9.44	& \textbf{10.00}	& \multicolumn{1}{|c}{72.9}\\ \hline

Saarplan  & 0.00	& 0.00	& \textbf{9.88}	& 7.99	& 0.00	& 10.00	& 0.00	& 0.00	& \textbf{10.00}	& 0.95	& \multicolumn{1}{|c}{38.8}\\ \hline

FD (A*) & 0.00	& 0.00	& 7.25	& \textbf{9.88}	& 0.00	& 10.00	& 0.00	& 0.00	& 10.00	& 0.78	& \multicolumn{1}{|c}{37.9}\\ \hline

FD (LAMA) & 0.00	& 0.00	& 4.46	& 4.47	& 0.00	& 9.71	& 0.00	& 0.00	& 9.75	& 0.00	& \multicolumn{1}{|c}{28.4}\\ \hline

Madagascar & 0.68	& 0.00	& 1.62	& 0.00	& 0.00	& 1.66	& 0.00	& 0.97	& 0.54	& 3.44	& \multicolumn{1}{|c}{8.9} \\ \hline

\end{tabular}
\caption{Results for agile score.}
\label{agile}
\end{table}

\begin{table}
\makebox[\textwidth]{%
\begin{tabular}{|l|ccccc}
\cline{1-1}
\textbf{Mean times (s)} & \multicolumn{1}{l}{\textbf{DUAL-BFWS}} & \multicolumn{1}{l}{\textbf{Saarplan}} & \multicolumn{1}{l}{\textbf{FD (A*)}} & \multicolumn{1}{l}{\textbf{FD (LAMA)}} & \multicolumn{1}{l}{\textbf{Madagascar}} \\ \hline
bait                & \multicolumn{1}{c|}{119.26}            & \multicolumn{1}{c|}{-}                & \multicolumn{1}{c|}{-}               & \multicolumn{1}{c|}{-}                 & \multicolumn{1}{c|}{8.94}               \\ \hline
boulderdash         & \multicolumn{1}{c|}{22.91}             & \multicolumn{1}{c|}{-}                & \multicolumn{1}{c|}{-}               & \multicolumn{1}{c|}{-}                 & \multicolumn{1}{c|}{-}                  \\ \hline
butterflies         & \multicolumn{1}{c|}{4.48}              & \multicolumn{1}{c|}{0.54}             & \multicolumn{1}{c|}{71.35}           & \multicolumn{1}{c|}{103.42}            & \multicolumn{1}{c|}{262.02}             \\ \hline
cakybaky            & \multicolumn{1}{c|}{22.02}             & \multicolumn{1}{c|}{4.02}             & \multicolumn{1}{c|}{1.08}            & \multicolumn{1}{c|}{43.92}             & \multicolumn{1}{c|}{-}                  \\ \hline
chipschallenge      & \multicolumn{1}{c|}{34.18}             & \multicolumn{1}{c|}{-}                & \multicolumn{1}{c|}{-}               & \multicolumn{1}{c|}{-}                 & \multicolumn{1}{c|}{-}                  \\ \hline
iceandfire          & \multicolumn{1}{c|}{0.46}              & \multicolumn{1}{c|}{0.12}             & \multicolumn{1}{c|}{0.12}            & \multicolumn{1}{c|}{1.22}              & \multicolumn{1}{c|}{369.77}             \\ \hline
simplified-bould.   & \multicolumn{1}{c|}{11.88}             & \multicolumn{1}{c|}{-}                & \multicolumn{1}{c|}{-}               & \multicolumn{1}{c|}{-}                 & \multicolumn{1}{c|}{-}                  \\ \hline
sokoban             & \multicolumn{1}{c|}{78.78}             & \multicolumn{1}{c|}{-}                & \multicolumn{1}{c|}{-}               & \multicolumn{1}{c|}{-}                 & \multicolumn{1}{c|}{1.22}               \\ \hline
zelda               & \multicolumn{1}{c|}{1.48}              & \multicolumn{1}{c|}{0.08}             & \multicolumn{1}{c|}{0.11}            & \multicolumn{1}{c|}{1.18}              & \multicolumn{1}{c|}{574.66}             \\ \hline
zenpuzzle           & \multicolumn{1}{c|}{0.22}              & \multicolumn{1}{c|}{1.37}             & \multicolumn{1}{c|}{4.34}            & \multicolumn{1}{c|}{-}                 & \multicolumn{1}{c|}{149.69}             \\ \hline
\end{tabular}}
\caption{Mean time for each game.}
\label{mean_game_times}
\end{table}

\begin{table}
\makebox[\textwidth]{%
\begin{tabular}{|l|ccccc}
\cline{1-1}
\textbf{Times STD (s)} & \multicolumn{1}{l}{\textbf{DUAL-BFWS}} & \multicolumn{1}{l}{\textbf{Saarplan}} & \multicolumn{1}{l}{\textbf{FD (A*)}} & \multicolumn{1}{l}{\textbf{FD (LAMA)}} & \multicolumn{1}{l}{\textbf{Madagascar}} \\ \hline
bait               & \multicolumn{1}{c|}{210.55}            & \multicolumn{1}{c|}{-}                & \multicolumn{1}{c|}{-}               & \multicolumn{1}{c|}{-}                 & \multicolumn{1}{c|}{0.00}               \\ \hline
boulderdash        & \multicolumn{1}{c|}{13.84}             & \multicolumn{1}{c|}{-}                & \multicolumn{1}{c|}{-}               & \multicolumn{1}{c|}{-}                 & \multicolumn{1}{c|}{-}                  \\ \hline
butterflies        & \multicolumn{1}{c|}{3.57}              & \multicolumn{1}{c|}{0.56}             & \multicolumn{1}{c|}{123.87}          & \multicolumn{1}{c|}{219.50}            & \multicolumn{1}{c|}{126.47}             \\ \hline
cakybaky           & \multicolumn{1}{c|}{10.20}             & \multicolumn{1}{c|}{0.89}             & \multicolumn{1}{c|}{0.12}            & \multicolumn{1}{c|}{9.91}              & \multicolumn{1}{c|}{-}                  \\ \hline
chipschallenge     & \multicolumn{1}{c|}{65.61}             & \multicolumn{1}{c|}{-}                & \multicolumn{1}{c|}{-}               & \multicolumn{1}{c|}{-}                 & \multicolumn{1}{c|}{-}                  \\ \hline
iceandfire         & \multicolumn{1}{c|}{0.11}              & \multicolumn{1}{c|}{0.04}             & \multicolumn{1}{c|}{0.03}            & \multicolumn{1}{c|}{0.39}              & \multicolumn{1}{c|}{237.16}             \\ \hline
simplified-bould.  & \multicolumn{1}{c|}{4.62}              & \multicolumn{1}{c|}{-}                & \multicolumn{1}{c|}{-}               & \multicolumn{1}{c|}{-}                 & \multicolumn{1}{c|}{-}                  \\ \hline
sokoban            & \multicolumn{1}{c|}{98.91}             & \multicolumn{1}{c|}{-}                & \multicolumn{1}{c|}{-}               & \multicolumn{1}{c|}{-}                 & \multicolumn{1}{c|}{0.00}               \\ \hline
zelda              & \multicolumn{1}{c|}{0.58}              & \multicolumn{1}{c|}{0.03}             & \multicolumn{1}{c|}{0.02}            & \multicolumn{1}{c|}{0.12}              & \multicolumn{1}{c|}{258.59}             \\ \hline
zenpuzzle          & \multicolumn{1}{c|}{0.28}              & \multicolumn{1}{c|}{0.00}             & \multicolumn{1}{c|}{0.00}            & \multicolumn{1}{c|}{-}                 & \multicolumn{1}{c|}{231.24}             \\ \hline
\end{tabular}}
\caption{Standard deviation for each game.}
\label{std_game_times}
\end{table}


Tables \ref{coverage}, \ref{satisficing} and \ref{agile} show planners results in each track and Tables \ref{mean_game_times} and \ref{std_game_times} the mean time and standard deviation for every game. From them we can see that coverage results are notoriously poor in most cases. Although the PDDL domains and problems are optimised to reduce the number of instances and actions, only seven games are solved by more than one planner.
    
This is mostly caused by the time consuming process of grounding\footnote{Transforming the problem from a lifted representation into a equivalent propositional one, in order to accelerate the planner search.}, exploding in time and memory. DUAL-BFWS seems to not be affected due to the initial best-width search it launches, getting a provisional plan early in the execution. However, these plans are non-optimal, and in almost every game where FD based planners resist the grounding process they found a better plan at a higher speed.

Therefore, the results show that planner behaviour should be improved in problems with many instances as the gain of speed does not justify the negative impact in coverage caused by the grounding process.


\section{Conclusions}\label{conclusions}

In this paper we have presented an automated knowledge engineering process to generate a huge variety of planning domains for video games\footnote{Sources and further detailed experiments are publicly available at \url{https://github.com/IgnacioVellido/VGDL-PDDL}}. We have shown experimental results on 10 video games represented in the video game description language VGDL, and potentially we are able to automatically generate domains for any game defined in this language, including all the video game repertory of the GVGAI framework (at present, more than one hundred). The process requires to represent behaviour patterns of avatars and objects in a knowledge base which are latter used as input by a compiling process, generating domains ready to be used by a classical planner. It has also been shown that this process, combined with a reactive component in a planning-based agent, can act and solve any of these games with only the prior knowledge of the game dynamics.

While the abstraction and definition of templates requires some knowledge of the game engine as well as the video game description language, we estimate that we can save weeks of work for a knowledge engineer devoted to design planning-based deliberative agents to act on these video games.

Furthermore, we performed benchmarks with state-of-the-art planners and showed that video game based domains offer diverse, understandable and challenging cases of study that can be of interest for the International Planning Community to consider future improvements. The experiments showed notorious bad results for problems with many instances, mainly due to the grounding process.

As far as the methodology goes, we feel that this project still has some points that can be improved upon and be regarded as future work:

\begin{itemize}
    \item \textit{Representation of non-determinism}: This kind of situations appear frequently not only in video games, but also in real life problems. In some cases a continuous planning strategy integrating replanning with reactive behaviour can be enough to face this problem, although if we aim to generate a flexible and self-contained domain, we think that the best approach is the introduction of non-deterministic planning techniques \citep{Kuter}.

    Other interesting approaches could consider fully observable non-deterministic (FOND) techniques like \citep{fond,Muise_McIlraith_Belle_2014}. However, further changes should be included in the domains to describe the non-deterministic characteristics of the games.
    
    \item \textit{Reduce grounding effects}: Although a human player can easily understand all the information in a level exact knowledge of every instance is frequently not of interest to solve the game. The overhead in time and memory caused by the many possible instantiations in these domains make the actual use of planning counterproductive. From a planning point of view, we believe that lifted or partial grounding strategies, like \citep{lifted, partial_grounding}, can make notorious improvements.
    
    \item \textit{Subgoal selection}: As we already noted throughout the paper, not only finding a plan can be excessively time consuming in these domains but the dynamic events of the games also makes large plans useless.
    
    We believe that a subobjective selection process can improve planner speed and plan optimisation. However, this is not an easy task, because we are no longer considering an unique parsing process. It is necessary to analyse the game and comprehend it. At present we are addressing the automated discovery of game strategies with the proposed architecture in \citep{Carlos}.
\end{itemize}


\bibliographystyle{kluwer}
\bibliography{References}

\end{document}